\documentclass{article}

\usepackage{arxiv}

\usepackage[utf8]{inputenc} 
\usepackage[T1]{fontenc}    
\usepackage{hyperref}       
\usepackage{url}            
\usepackage{booktabs}       
\usepackage{amsfonts}       
\usepackage{nicefrac}       
\usepackage{microtype}      
\usepackage{lipsum}		
\usepackage{graphicx}
\usepackage{natbib}
\usepackage{doi}

\usepackage{adjustbox}
\usepackage{makecell}
\usepackage{verbatim}
\usepackage{tabularx}
\usepackage{amsmath}

\usepackage{listings}
\usepackage{booktabs}
\usepackage{xcolor}
\usepackage{hhline}
\usepackage{multirow}
\usepackage{subcaption}

\title{MiniGPT-Pancreas: Multimodal Large Language Model for Pancreas Cancer Classification and Detection}

\author{ \href{https://orcid.org/0000-0002-3365-580X}{\includegraphics[scale=0.09]{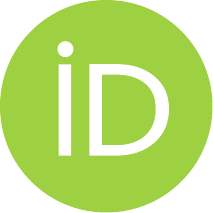}\hspace{1mm}Andrea Moglia} \\
	Department of Electronics, \\Information, and Bioengineering\\
	Polytechnic University of Milan\\
	Milan, 20133, Italy \\
	\texttt{andrea.moglia@polimi.it} \\
	\And
	{\includegraphics[scale=0.0]{orcid.pdf}\hspace{1mm}Elia Clement Nastasio} \\
	Department of Electronics, \\Information, and Bioengineering\\
	Polytechnic University of Milan\\
	Milan, 20133, Italy \\
	\texttt{elia.clement.nastasio@gmail.com} \\
        \And
        \href{https://orcid.org/0000-0002-6276-6314}{\includegraphics[scale=0.09]{orcid.pdf}\hspace{1mm}Luca Mainardi} \\
	Department of Electronics, \\Information, and Bioengineering\\
	Polytechnic University of Milan\\
	Milan, 20133, Italy \\
	\texttt{luca.mainardi@polimi.it} \\
        \And
        \href{https://orcid.org/0000-0003-3995-8673}{\includegraphics[scale=0.09]{orcid.pdf}\hspace{1mm}Pietro Cerveri} \\
	Department of Industrial, \\and Information       Engineering\\
	University of Pavia\\
	Pavia, 27100, Italy \\
	\texttt{pietro.cerveri@unipv.it} \\
        Department of Electronics, \\Information, and Bioengineering\\
	Polytechnic University of Milan\\
	Milan, 20133, Italy \\
	\texttt{pietro.cerveri@polimi.it} \\
}

\renewcommand{\shorttitle}{MiniGPT-Pancreas: Multimodal Large Language Model for Pancreas Cancer Diagnosis}

\hypersetup{
pdftitle={MiniGPT-Pancreas: Multimodal Large Language Model for Pancreas Cancer Classification and Detection},
pdfsubject={cs.AI, cs.CV},
pdfauthor={Andrea Moglia, Elia Clement Nastasio, Luca Mainardi, Pietro Cerveri},
pdfkeywords={Artificial intelligence pancreas segmentation, Deep learning pancreas segmentation, Pancreas segmentation, Pancreas tumor segmentation},
}

\lstdefinelanguage{json}{
    basicstyle=\ttfamily,
    numbers=left,
    numberstyle=\tiny\color{gray},
    stepnumber=1,
    numbersep=8pt,
    showstringspaces=false,
    breaklines=true,
    frame=single,
    backgroundcolor=\color{lightgray},
    literate=
     *{0}{{{\color{blue}0}}}{1}
      {1}{{{\color{blue}1}}}{1}
      {2}{{{\color{blue}2}}}{1}
      {3}{{{\color{blue}3}}}{1}
      {4}{{{\color{blue}4}}}{1}
      {5}{{{\color{blue}5}}}{1}
      {6}{{{\color{blue}6}}}{1}
      {7}{{{\color{blue}7}}}{1}
      {8}{{{\color{blue}8}}}{1}
      {9}{{{\color{blue}9}}}{1}
      {:}{{{\color{red}:}}}{1}
      {,}{{{\color{red},}}}{1}
      {\{}{{{\color{black}{\{}}}}{1}
      {\}}{{{\color{black}{\}}}}}{1}
      {[}{{{\color{black}{[}}}}{1}
      {]}{{{\color{black}{]}}}}{1},
}

\begin{document}
\maketitle

\begin{abstract}
\textit{Problem:} Pancreas radiological imaging is challenging due to the small size, blurred boundaries, and variability of shape and position of the organ among patients. \textit{Goal:} In this work we present MiniGPT-Pancreas, a Multimodal Large Language Model (MLLM), as an interactive chatbot to support clinicians in pancreas cancer diagnosis by integrating visual and textual information. \textit{Methods:} MiniGPT-v2, a general-purpose MLLM, was fine-tuned in a cascaded way for pancreas detection, tumor classification, and tumor detection with multimodal prompts combining questions and computed tomography scans from the National Institute of Health (NIH), and Medical Segmentation Decathlon (MSD) datasets. The AbdomenCT-1k dataset was used to detect the liver, spleen, kidney, and pancreas. \textit{Results:} MiniGPT-Pancreas achieved an Intersection over Union (IoU) of 0.595 and 0.550 for the detection of pancreas on NIH and MSD datasets, respectively. For the pancreas cancer classification task on the MSD dataset, accuracy, precision, and recall were 0.876, 0.874, and 0.878, respectively. When evaluating MiniGPT-Pancreas on the AbdomenCT-1k dataset for multi-organ detection, the IoU was 0.8399 for the liver, 0.722 for the kidney, 0.705 for the spleen, and 0.497 for the pancreas. For the pancreas tumor detection task, the IoU score was 0.168 on the MSD dataset. \textit{Conclusions}: MiniGPT-Pancreas represents a promising solution to support clinicians in the classification of pancreas images with pancreas tumors. Future research is needed to improve the score on the detection task, especially for pancreas tumors. 
\end{abstract}


\section{Introduction}

Pancreas cancer is notoriously deadly, with a five-year survival rate of 13\% in the United States, making it the lowest one among all cancer types~\citep{siegel2024cancer}. Its lethality is compounded by the fact that surgical resection is the only potentially curative treatment, but only a small fraction of patients are eligible for surgery due to the advanced stage at diagnosis time. Even among those who undergo surgery, recurrence rates are high, and long-term survival remains low. For this reason, an early diagnosis of pancreatic cancer is crucial. Diagnosis involves clinical assessment, laboratory testing, and advanced imaging techniques like computed tomography (CT). Analysis of pancreas radiological imaging is demanding for several reasons. First, the pancreas has an irregular shape and can be easily deformed. Its shape, size, aspect ratio, position, and orientation inside the abdomen cavity vary largely among subjects \citep{dai2023td, man2019deep}. Second, the organ is very small, and its volume is generally less than 0.5\% of an entire CT \citep{man2019deep}. Third, there is low contrast in the CT slices between pancreas boundaries and the other abdominal structures.
Unfortunately, tumors smaller than 2~cm often evade detection by CT, emphasizing the need for new methods to support
clinicians~\citep{dewitt2006comparison}.

Artificial intelligence (AI) models have made great strides in screening, diagnosis, treatment guidance, and prognosis prediction in oncology over the past few years~\citep{kann2021artificial}. 
However, AI models face challenges in pancreas cancer imaging as traditional approaches. For instance, state-of-the-art AI architectures hardly reach an accuracy of 0.60 on tumor segmentation on commonly public datasets~\citep{moglia2024deep}. This is lower compared to cancers in other organs, such as the liver, where an accuracy of up to 0.70 has been reported on similar datasets~\citep{bilic2023liver}. More recently, large language models (LLMs), a type of AI models pre-trained on massive data using unsupervised learning and fine-tuned on specific downstream tasks, were applied to healthcare. They have demonstrated their capability to pass medical examinations, requested by national boards in different countries to obtain the license for clinical practice~\citep{moglia2024large}. However, those LLMs are unimodal. As such, they are not suitable for patient diagnosis and treatment, which mandates a holistic approach combining different sources of data, e.g., medical history, electronic health records, laboratory tests, and radiology results~\citep{alsaad2024multimodal}.
By combining multimodal data, e.g., text and image, multimodal large language models (MLLMs) have the potential to revolutionize healthcare, for instance, by interpreting patient history or summarizing findings. Moreover, the ability of LLMs, and consequently MLLMs, to process natural language queries and provide straightforward answers could streamline clinical decision-making processes. Furthermore, the capabilities of MLLMs-based chatbots could prove instrumental in assisting young medical practitioners when needing a second opinion from a more experienced colleague. Thus, MLLM-based medical chatbots may relieve the burden on the healthcare system, thus improving efficiency~\citep{lin2023medical}.

LLaVA-Med, Med-Flamingo, and Med-PaLM M represented initial instances of generalist MLLMs accepting text and images from different modalities, e.g., CT, X-ray, and Magnetic Resonance Imaging (MRI), and generating text as output for tasks in radiology, dermatology, and pathology~\citep{li2024llava, moor2023med, he2024pefomed}. As such, they were specialized for the visual question answering (VQA) task. However, they were not suitable for additional tasks, e.g. organ and lesion detection, hence limiting their use.
Unfortunately, developing and training an MLLM for VQA and other clinical tasks is challenging for several reasons. First, the distinction of information within medical domains is much more subtle and fine-grained than in natural images. Second, curating a large-scale dataset for medical applications requires expert annotation, which is expensive in terms of time and costs. 
Third, collecting patients' data, e.g. medical images, requires privacy protection to comply with security laws, like the European Union General Data Protection Regulation (GDPR). 
A possible solution could be to fine-tune an existing MLLM for a specific task.
MiniGPT-v2 was proposed as a general-purpose MLLM for different tasks, such as image captioning, VQA, referring expression comprehension (REC), i.e. identification and detection of a specific object in
an image that is referred to by a natural language
expression~\citep{chen2023minigpt}. 
MiniGPT-Med, a fine-tuned example of MiniGPT-v2, was developed for several tasks in lung nodule diagnosis, e.g., detection, in addition to VQA~\citep{alkhaldi2024minigpt}.
At present, there is no published study on MLLMs for pancreas diagnosis support, e.g. for localizing the organ inside a CT scan after receiving a text prompt. 
To address this need, we propose MiniGPT-Pancreas, an MLLM for pancreas detection, tumor detection, and classification tasks. It is based on a cascade fine-tuning of MiniGPT-v2 on the National Institute of Health (NIH), and the Medical Segmentation Decathlon (MSD), two publicly available datasets~\citep{roth2015deeporgan, simpson2019large}. It was finally tested on the AbdomenCT-1k dataset for the multi-organ detection task~\citep{ma2021abdomenct}. The main contributions of this work are the following:

\begin{enumerate}
    \item  we introduce MiniGPT-Pancreas, the first MLLM for pancreas diagnostic support allowing interactive dialogue with users in a conversational ChatGPT style;
    \item we performed a cascade fine-tuning and extensive evaluation of MiniGPT-Pancreas for pancreas detection, tumor classification, and detection. More specifically, for each new task, MiniGPT-Pancreas was fine-tuned on the checkpoint of the previous task;
    \item we showed that MiniGPT-Pancreas outperformed MiniGPT-v2 by a large margin in all the tasks. We also assessed it on the detection of the pancreas, liver, spleen, and kidney on the AbdomenCT-1k dataset.
\end{enumerate}

The remaining of this manuscript is structured as follows: in Section \ref{sec:related_work} we report the related work on MLLM for medicine. In Section \ref{sec:materials_methods}, we describe the architecture of MiniGPT-Pancreas, the prompt template, the preprocessing of CT volumes, the training pipeline, the evaluation metrics, and the implementation of the training sessions. In Section \ref{sec:results} we report the results on pancreas detection, tumor classification, and detection task; we also present the results on the multi-organ detection task. In Section \ref{sec:discussion} we discuss our findings, along with limitations and future developments. Section \ref{sec:conclusion} ends the manuscript. 

Our model and code have been made publicly available at: \href{https://github.com/elianastasio/MiniGPT-Pancreas}{https://github.com/elianastasio/MiniGPT-Pancreas}

\section{Related Work}
\label{sec:related_work}
Traditional medical VQA tasks were approached by AI models as classification problems, where the model was trained to categorize image-text representations into a predefined set of answers. However, this approach was limited by its reliance on a fixed set of candidate answers, restricting its utility for open-ended questions that are common in medical diagnostics \citep{he2024pefomed}. Recent advancements have led to the development of several MLLMs generating open-ended responses rather than selecting from pre-defined answer choices in the healthcare domain \citep{he2024pefomed}.

BiomedGPT was the first open-source MLLM for healthcare, capable of performing VQA, report generation, and summarization \citep{zhang2023biomedgpt}. It was pre-trained on 14 freely available datasets and fine-tuned on 19 datasets for VQA and image captioning tasks. Its architecture consisted of an encoder-decoder LLM and a convolutional neural network as the visual encoder. BiomedGPT outperformed the state-of-the-art (SOTA) on the image classification task \citep{zhang2023biomedgpt}.

Med-PaLM M was proposed by Google (Mountain View, CA, United States) for radiological image classification, VQA, report generation, and summarization \citep{tu2024towards}. It was based on the Pathways Language Model (PaLM) decoder transformer and a vision transformer as the visual encoder. Med-PaLM M was developed by fine-tuning PaLM-E on 12 open-source datasets and 14 tasks. It scored better than SOTA on VQA, report generation, and image classification \citep{tu2024towards}.

XrayGPT was designed as a conversational MLLM for analyzing chest radiographs and generating summaries from radiological reports \citep{thawakar2024xraygpt}. It consisted of a visual encoder, an LLM, and a linear transformation layer to align visual features with the LLM. During training, only the linear transformation layer was updated. Med-CLIP and Vicuna were employed as visual and LLM, respectively. It was trained in two stages using MIMIC-CRX and OpenI datasets, both containing X-ray images and reports \citep{johnson2019mimic, demner2016preparing}.

LLaVA-Med, an extension of LLaVA (Large Language and Vision Assistant), was specifically designed for the medical domain \citep{li2024llava}. It was fine-tuned using specialized datasets, like PMC-15M, containing 15 million biomedical image-text pairs derived from PubMed Central, and filtered using GPT-4. The development of LLaVA-Med followed a two-stage learning process. In the first stage, the model was trained using image-text pairs to predict the original image caption to ensure that the visual and textual concepts in the biomedical domain were accurately aligned. The second stage involved instruction-tuning, where the model learned to follow more complex and open-ended medical instructions. One of the key innovations in LLaVA-Med was the novel data generation pipeline, leveraging GPT-4 to create diverse instruction-following instances from the PMC-15M dataset without the need of manual annotation. This approach ensured that the model was trained on a wide range of medical concepts, improving its zero-shot performance (the ability of a model to perform a task without having been explicitly trained on any examples of that task) for medical VQA. Experiments showed that LLaVA-Med outperformed its general-domain model, LLaVA, in various benchmark datasets such as VQA-RAD, Slake, and PathVQA, often achieving state-of-the-art results\citep{lau2018dataset, he2020pathvqa}.

Med-Flamingo, a vision-language model based on OpenFlamingo-9B for the medical domain, was developed to handle complex, interleaved image-text data for few-shot learning (the ability of an MLLM/LLM to make a prediction after receiving a prompt with few examples) \citep{moor2023med, awadalla2023openflamingo}. Trained on several medical textbooks and PubMedCentral OpenAccess (PMC-OA) data, Med-Flamingo addressed the limitations of previous multimodal medical models such as ChexZero and BiomedCLIP, which lacked the ability to perform in-context learning with few-shot examples \citep{tiu2022expert, zhang2023biomedclip}.

PeFoMed (Parameter Efficient Fine-tuning of Multimodal Large Language Models for Medical Imaging) was proposed as a fine-tuned MiniGPT-v2 using low-rank adaptation (LoRA) to reduce computational resources \citep{he2024pefomed, hu2021lora}. PeFoMed fine-tuning process involved a two-stage approach, keeping the vision encoder and language model frozen, while updating only the linear projection and the LoRA layers in both stages. In the first one, the model was fine-tuned using ROCO, a dataset coupling medical images and their corresponding textual descriptions to adapt the model to understand and generate medical text based on visual inputs \citep{ruckert2024rocov2}.  In the second stage, the model was fine-tuned on the VQA-RAD dataset \citep{lau2018dataset}. Remarkably, PeFoMed outperformed GPT-4 by a significant margin on the VQA-RAD dataset, particularly for open-ended questions. PeFoMed outperformed LLaVA-Med on the VQA task on the VQA-RAD dataset on both open-ended and closed-ended questions, achieving an overall score of 81.9\% of correct answers vs. 75.8\%~\citep{he2024pefomed}.

CheXagent, an MLLM specifically focused on chest X-rays, was trained in four stages on 28 public datasets for eight tasks like VQA, disease classification, and abnormality detection~\citep{chen2024chexagent}. It was based on Mistral-7B-v01 LLM and BLIP-2 visual encoder \citep{li2023blip}. In the first training stage, CheXagent was trained with different sources like PubMed Central abstracts, and medical terms from Wikipedia. In the second, BLIP-2 was trained for image-text contrastive learning and image captioning objectives using image-text pairs datasets such as MIMIC-CXR. In the third, a model to bridge the LLM and visual encoder was trained using the image captioning objective. In the last stage, CheXagent was trained on a variety of tasks. CheXagent outperformed XrayGPT, Med-Flamingo, RadFM, and LLaVA-Med on several tasks including QVA, and disease classification \citep{chen2024chexagent}.

Radiology Foundation Model (RadFM), was developed starting from pre-training on the Medical Multi-modal Dataset (MedMD), consisting of 16 million 2D and 3D images collected from different datasets and accompanied with textual descriptions, such as radiology reports, and visual-language instructions \citep{wu2023towards}. It was then fine-tuned on three million multi-modal samples of MedMD with only radiologic images. RadFM consisted of a 3D visual transformer, and MedLLaMA-13B, a fine-tuned version of LLaMA-13B. RadFM outperformed MedFlamingo, and GPT-4V on VQA, report generation, and rationale diagnosis tasks \citep{wu2023towards}.

MiniGPT-Med, a fine-tuned version of MiniGPT-v2, was designed for medical report generation, VQA, and disease identification on lung nodules \citep{alkhaldi2024minigpt}. It was based on EVA visual encoder and LLaMA2-7B-chat as LLM \citep{fang2023eva, touvron2023llama}. On the medical report generation task, MiniGPT-Med outperformed MedFlamingo, LLaVA-Med, RadFM, XrayGPT, CheXagent, and MiniGPT-v2. On disease detection, it scored better than Bio-ViL-T, MedKLIP, and MiniGPT-v2 \citep{bannur2023learning, wu2023medklip}. On VQA, it outperformed MedFlamingo and MiniGPT-v2 \citep{moor2023med}.

\section{Material and Methods}
\label{sec:materials_methods}

\begin{figure*}[!t]
	\centering
	\includegraphics[width=0.8\linewidth]{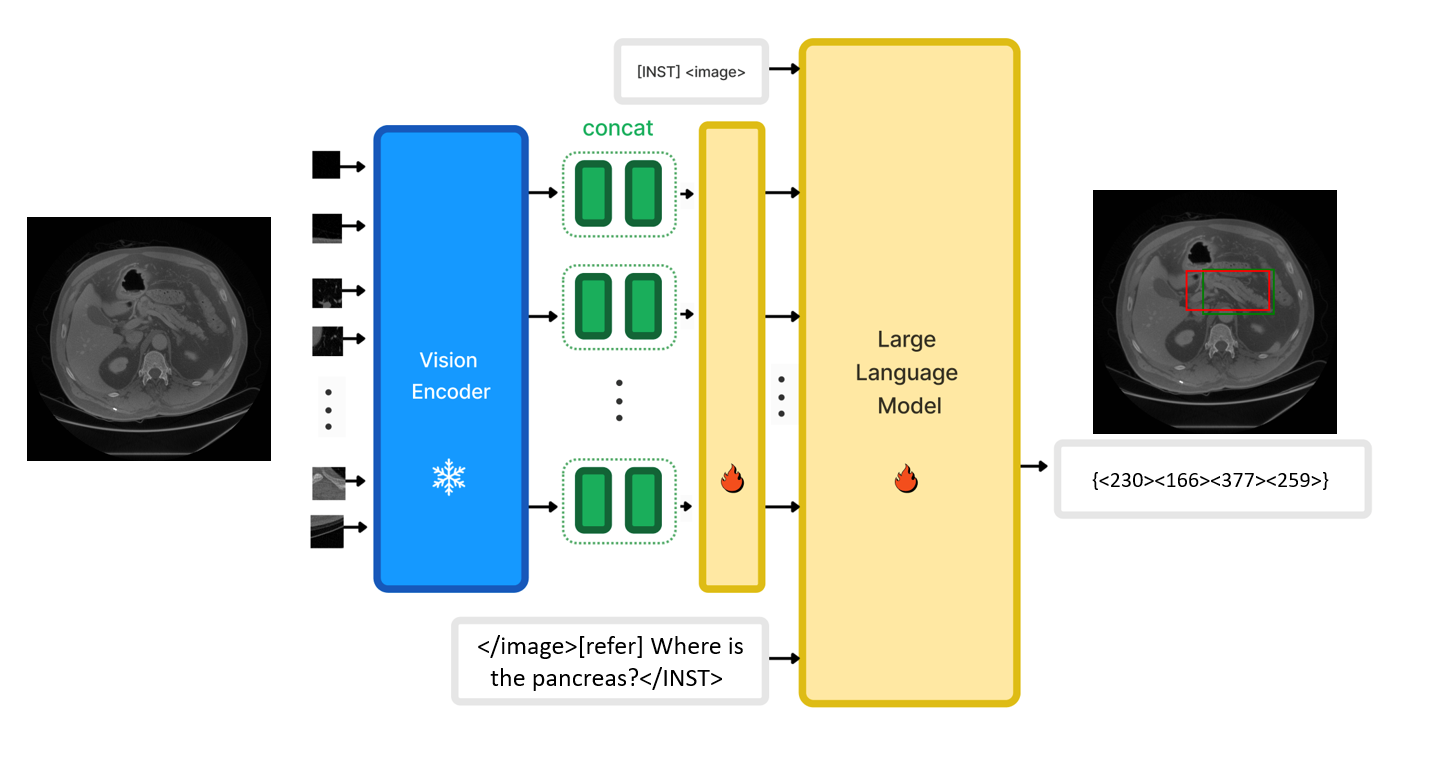}
	\caption{Architecture of MiniGPT-Pancreas consisting of a vision encoder, a linear projection layer,
and an LLM. The visual encoder is fed with each CT slice and one text prompt, e.g. a question. The model then generates text as output, e.g. an answer to a question or the coordinates of a bounding box as text. Adapted from \citep{alkhaldi2024minigpt}.}
	\label{fig:minigpt-v2_architecture}
\end{figure*}

\subsection{Model Architecture}
\label{subsec:minigpt-v2_architecture}
In this work, MiniGPT-v2 was adopted as the base MLLM for MiniGPT-Pancreas. It combines a visual encoder, a projection layer, and a pre-trained LLM, as depicted in Fig. \ref{fig:minigpt-v2_architecture}~\citep{chen2023minigpt}. 

\textbf{Visual encoder:} EVA, a vanilla vision transformer (ViT) with 1 billion parameters, was used as a visual encoder \citep{fang2023eva}. 

\textbf{Projection layer:} A linear projection layer was placed between the visual encoder and the LLM to serve as a bridge between the visual and textual modalities. Its role was to translate the extracted features from the ViT into input tokens compatible with the LLM format. By aligning the dimensionality of the ViT output with that of the language tokens, the projection layer ensured that the LLM could process the visual information in a manner consistent with textual data. This transformation was crucial for the seamless integration of visual and textual inputs, allowing the LLM to interpret and generate contextually appropriate responses based on the visual cues provided
by the ViT.

\textbf{Pre-trained LLM:} LLaMA-2-chat-7b-hf, a seven billion parameter model, belonging to the open-source LLaMA-2 ecosystem, was employed as LLM \citep{touvron2023llama}. The \texttt{chat} suffix refers to the additional fine-tuning process that the LLM underwent, optimizing it for dialogue use cases. The \texttt{hf} suffix stands for human feedback and refers to reinforcement learning from human feedback, a technique where humans evaluated the model outputs, e.g., responses in a conversation, and provided feedback on their quality. This feedback was then used to adjust the model's behavior, making it more aligned with human preferences. 

During the training of MiniGPT-v2, the visual encoder was kept frozen, while the projection layer and the LLM were updated \citep{chen2023minigpt}.

\subsection{Prompt Template}
\label{subsec:prompt-template}
A distinctive feature introduced by MiniGPT-v2 was a multitask instruction template to generate the most suitable format for the output of the various tasks, by inserting a task-specific token, e.g. \texttt{vqa} for VQA and \texttt{refer} for REC tasks, respectively. The instruction template had the following format:

\noindent
\begin{flushleft}\textit{[INST] \textless Img \textgreater \textless ImageFeature \textgreater \textless/Img\textgreater [Task Identifier] Instruction [/INST]}
\end{flushleft}

The template consisted of the visual features extracted by the ViT, the task identifier, and the instruction in natural language.
Without the task identifier, the MLLM could fail to identify the current instruction. For instance, when the model was asked to detect the pancreas in a CT slice, it could generate an answer in natural language (e.g., "between the stomach and the spine") instead of outputting a bounding box as expected. 

The instruction component represented the natural language part of the prompt (e.g., 'Where is the pancreas?'). During training, an instruction was randomly selected from a list of available candidates, helping the model to generalize better and to reduce the risk of overfitting to a specific text sequence. The following candidate instructions, preceded by the \texttt{refer} task identifier, were designed for the pancreas detection task:
\begin{itemize}
    \item "\texttt{[refer]} Give me the location of the pancreas"
    \item "\texttt{[refer]} Give me the position of the pancreas"
    \item "\texttt{[refer]} Where is the pancreas?"
    \item "\texttt{[refer]} Where is located the pancreas in this image?"
    \item "\texttt{[refer]} Which is the position of the pancreas?"
    \item "\texttt{[refer]} From this image, tell me the location of the pancreas"
    \item "\texttt{[refer]} Could you tell me the location of the pancreas?"
    \item "\texttt{[refer]} Where can I locate the pancreas?"
\end{itemize}

Similar instructions were designed for pancreas tumor detection. For the task of tumor classification, the following \texttt{YES/NO} questions were designed (preceded by the \texttt{vqa} task identifier):

\begin{itemize}
    \item "\texttt{[vqa]} Does the pancreas in the image present a tumor?"
    \item "\texttt{[vqa]} Is there a tumor in the pancreas shown in the image?"
    \item "\texttt{[vqa]} Can you see a tumor in the pancreas in this picture?"
    \item "\texttt{[vqa]} Does the image show a tumor in the pancreas?"
    \item "\texttt{[vqa]} Is a pancreatic tumor visible in the image?"
    \item "\texttt{[vqa]} Is the pancreas in this image showing signs of a tumor?"
    \item "\texttt{[vqa]} Is the pancreas in the image affected by a tumor?"
    \item "\texttt{[vqa]} Does the pancreas in the picture have a tumor?"
    
\end{itemize}

\subsection{Bounding Boxes Representation}
\label{subsec:bbox-representation}
For grounding tasks based on localization of objects like detection, the bounding boxes were represented as numbers in natural language: 
\[{< X_{\text{left}} >< Y_{\text{top}} >< X_{\text{right}} >< Y_{\text{bottom}} >}\]
Each coordinate is an integer normalized on the [0, 100] range. Specifically, \(< X_{\text{left}} >< Y_{\text{top}} >\) represents the top-left corner, while \(< X_{\text{right}} >< Y_{\text{bottom}} >\) the bottom-right corner.

\subsection{Datasets}
\label{subsec:datasets}
The  NIH dataset includes 82 CTs in the arterial phase, with a 512 $\times$ 512 resolution, a number of slices ranging from 181 to 466, and a [1.5-2.5] slice thickness \citep{roth2015deeporgan}. The pancreas was manually labeled by a medical student and then verified by an experienced radiologist \citep{roth2015deeporgan}. The images of the MSD dataset were provided by the Memorial Sloan Kettering Cancer Center (New York, NY, United States). Of the 420 CT scans that were acquired in the venous phase, 281 included annotations of pancreas and tumors \citep{simpson2019large}. The AbdomenCT-1k is a multi-organ dataset with 1,112 CTs with annotations of liver, pancreas, kidneys, and spleen \citep{ma2021abdomenct}. A cohort of 15 junior annotators (one to five years of experience) used ITK-SNAP to manually segment the organs under the supervision of two board-certified radiologists. Then, one senior radiologist with more than 10 years of experience checked the annotations. After annotation, UNet models were trained to find the possible errors, which were double-checked by the senior radiologist \citep{ma2021abdomenct}. The AbdomenCT-1k dataset includes the MSD, and the NIH datasets, where the annotations of liver, spleen, and kidneys were added \citep{ma2021abdomenct}.

\subsection{Data Preprocessing}
\label{subsec:data_preprocessing}
In order to use the NIH, MSD, and AbdomenCT-1k datasets during training and testing, some preprocessing steps were performed to convert the 3D CT volumes into 2D images to align with the input dimensional requirements of the visual encoder of MiniGPT-Pancreas. 
Since each volume contained many slices, only those containing the pancreas were saved as 2D images in order to limit the computational cost. Additionally, before converting each slice to a \texttt{PNG} file, histogram cropping of 2\% and histogram equalization were performed to remove outliers, and improve image contrast, respectively. Concurrently with image preprocessing, a \texttt{JSON} file was generated, containing useful annotations for selecting the appropriate slices in the different training processes. An example is shown in Fig. \ref{fig:json_data}. In order to extract the bounding boxes of the anatomical regions of interest (either the pancreas parenchyma or the tumor), the top-left and bottom-right pixels in the annotated mask of the ground truth (GT) were computed. The definitions of all keys used in the \texttt{JSON} file are reported in Section \ref{app-sec:materials_methods} of the Appendix. Overall, the MSD and NIH datasets yielded 8,792 and 6,882 slices, respectively. In particular, in each slice, at least one pixel was labeled as pancreas.

\begin{figure}[h]
\centering
\begin{lstlisting}[language=json, numbers=none]
{
    "dataset": "MSD",
    "volume_name": "pancreas_228.nii.gz",
    "slice_id": 603,
    "slice_index": 52,
    "slice_count": 113,
    "pixels_pancreas": 804,
    "pancreas_pixels_ratio": 0.62,
    "max_pixels_pancreas": 1304,
    "pixels_tumor": 258,
    "tumor_pixels_ratio": 0.92,
    "max_pixels_tumor": 279,
    "bbox_pancreas": [
        196,
        235,
        237,
        260
    ],
    "pancreas_bbox_ratio": 0.39,
    "max_bbox_pancreas": 2652,
    "bbox_tumor": [
        220,
        238,
        237,
        255
    ],
    "tumor_bbox_ratio": 0.94,
    "max_bbox_tumor": 306,
    "width": 512,
    "height": 512
}
\end{lstlisting}
\caption{JSON data for a specific pancreas slice from the MSD dataset.}
\label{fig:json_data}
\end{figure}

\subsection{Training Pipeline}
\label{subsec:training_pipeline}
The following three training stages were performed.

\begin{enumerate}
    \item \textbf{Pancreas Detection:} During this step, the base checkpoint of MiniGPT-v2 was fine-tuned, yielding MiniGPT-Pancreas. In particular, the model was asked to generate a bounding box around the pancreas, using the \texttt{[refer]} task identifier. Both MSD and NIH datasets were employed. In order to evaluate the impact of using small or large GT bounding boxes containing the pancreas, 10 different training sessions were performed by choosing a different \texttt{pancreas\_bbox\_ratio} threshold (equivalent to the ratio between the pancreas bounding box and the largest pancreas bounding box in any slice of the same volume). The threshold ranged from 0\% to 90\%. This ensured that only the slices with at least that pancreas bounding box area ratio were selected.
    A train-test split of 80\%-20\% was chosen, based on the number of CT volumes.
    \item \textbf{Tumor Classification:} For this task, the MiniGPT-Pancreas model obtained from the previous task was fine-tuned to determine whether the images contained a pancreas tumor or not. In this case, the task identifier \texttt{[vqa]} was selected.
    \item \textbf{Tumor Detection:} For this task, the fine-tuned MiniGPT-Pancreas version for tumor classification was further fine-tuned to localize the pancreas tumors. The \texttt{[refer]} task identifier was employed as in the first stage. In this case, only the MSD dataset was used since it contained annotations of tumors, in contrast with the NIH dataset which included only healthy pancreas. A key difference from the first training stage was the significantly smaller size of the GT bounding boxes.
\end{enumerate}

\subsection{Evaluation Metrics}
\label{subsec:metrics}
For the detection task, the Intersection over Union (IoU) was adopted as an evaluation metric. Its formula is the following: 

\[
    \text{IoU} = \frac{|A \cap B|}{|A \cup B|} \quad
\]
where \(A\) represents the predicted bounding box and \(B\) represents the GT bounding box. The IoU measures the ratio of the intersection between the predicted and GT regions to their union.
Accuracy, precision, recall, and F1 score were used for the binary classification task.

\subsection{Implementation Details}
\label{subsec:implementation_details}

All training sessions were performed on a single NVIDIA A100 GPU with 40GB of memory. The model was initialized using the checkpoint of MiniGPT-v2. All weights of the ViT were kept frozen, while the linear projection layer was updated. The LLM was trained by applying LoRA to the \texttt{key} and \texttt{query} weights matrices of the attention layers, \(W_q\) and \(W_k\), with rank set to 64 and scaling factor $\alpha$ set to 16. This allowed us to reduce the number of trainable parameters of the LLaMA-2-chat-7b-hf to just 34 million, equivalent to approximately 0.5\% of the original model.
Image resolution was set to 448\(\times\)448 in both training and testing. AdamW was chosen as optimizer, in combination with a linear warm-up cosine scheduler. The initial learning rate was \(1e^-5\), while the warm-up and the minimum learning rate were both set to \(1e^-6\). Weight decay was set to 0.05. Cross-entropy was used as a loss function. Each fine-tuning process ran for 50 epochs, taking approximately 12 hours.

\subsection{Interactive Interface}
\label{subsec:implementation_details}

Once trained for the respective task, MiniGPT-Pancreas could be used for inference by submitting multimodal prompts to the native ChatGPT style web interface of MiniGPT-v2, designed with Gradio \footnote{https://www.gradio.app/}.
An example of a chat between the user and MiniGPT-Pancreas is depicted in Fig. \ref{app-fig:minigpt-v2_web_interface} of the Appendix.

\section{Results}
\label{sec:results}

\subsection{Pancreas Detection}
\label{subsec:pancreas_detection}
The number of slices included in the train and test splits of the NIH and MSD datasets are reported in Table \ref{app-tab:pancreas_detection_sizes} of the Appendix. For the pancreas detection task, several training sessions were performed by varying the \texttt{pancreas\_bbox\_ratio} threshold from 0\% to 90\%. Each training session lasted 10 epochs. 
The IoU scores on the NIH and MSD datasets, and their average weighted by the number of training and test slices are shown in Table \ref{app-tab:pancreas_detection_results} of the Appendix. A 60\% value of threshold was adopted as a trade-off between the IoU scores and the number of slices in the train and test split for each threshold level.

Table \ref{tab:epoch60} shows the results of the model trained for 50 epochs using the 60\% threshold. MiniGPT-Pancreas reached the highest IoU after 33 epochs on the NIH, and 21 epochs on the MSD dataset, respectively. The highest average was reached after 21 epochs (Table \ref{tab:epoch60}). These values were considerably higher than those obtained by the base MiniGPT-v2 before fine-tuning, i.e., 0.053, 0.043, and 0.047 on the NIH, MSD, and the average between the two datasets, respectively. Some examples of the pancreas detection task on the NIH and MSD datasets are shown in Fig. \ref{app-fig:NIH_pancreas_detection} and Fig. \ref{app-fig:MSD_pancreas_detection} of the Appendix.

\begin{table}[h]
    \centering
    \caption{Scores on IoU for the pancreas detection task with the threshold set at 60\%. The average was weighted by considering the number of slices. The best results are in bold.}
    \label{tab:epoch60}
    \resizebox{0.45\columnwidth}{!}{%
    \scriptsize
    \begin{tabular}{lccc}
        \hhline{====}
        \multicolumn{1}{l}{} & \multicolumn{1}{c}{\textbf{NIH}} & \multicolumn{1}{c}{\textbf{MSD}} & \multicolumn{1}{c}{\textbf{Average}} \\
        \cmidrule(){2-4}
        \textbf{Epoch} & \textbf{IoU} & \textbf{IoU} & \textbf{IoU} \\
        \midrule
        1  & 0.574 & 0.533 & 0.550 \\
        3  & 0.584 & 0.544 & 0.560 \\
        5  & 0.591 & 0.538 & 0.560 \\
        7  & 0.587 & 0.548 & 0.564 \\
        9  & 0.584 & 0.548 & 0.563 \\
        11 & 0.585 & 0.543 & 0.561 \\
        13 & 0.592 & 0.555 & 0.570 \\
        15 & 0.587 & 0.552 & 0.566 \\
        17 & 0.594 & 0.558 & 0.573 \\
        19 & 0.593 & 0.556 & 0.572 \\
        21 & 0.595 & \textbf{0.550} & \textbf{0.574} \\
        23 & 0.593 & 0.554 & 0.570 \\
        25 & 0.590 & 0.553 & 0.569 \\
        27 & 0.592 & 0.556 & 0.571 \\
        29 & 0.594 & 0.555 & 0.571 \\
        31 & 0.589 & 0.555 & 0.569 \\
        33 & \textbf{0.597} & 0.556 & 0.573 \\
        35 & 0.591 & 0.556 & 0.571 \\
        37 & 0.589 & 0.559 & 0.572 \\
        39 & 0.594 & 0.557 & 0.572\\
        41 & 0.596 & 0.556 & 0.573 \\
        43 & 0.595 & 0.555 & 0.572 \\
        45 & 0.595 & 0.556 & 0.572 \\
        47 & 0.593 & 0.557 & 0.572 \\
        49 & 0.594 & 0.556 & 0.572 \\
        \midrule
        MiniGPT-v2 & 0.053 & 0.043 & 0.047 \\
        \hhline{====}
    \end{tabular}
    }
\end{table}

\subsection{Pancreas Tumor Classification}
\label{subsec:tumor_classification}

For the tumor classification task, the MiniGPT-Pancreas model for pancreas detection was fine-tuned. In particular, we used the model achieving the highest average IoU score between NIH and MSD datasets, i.e., after 21 epochs using the 60\% threshold (Table \ref{tab:epoch60}). For the classification task, MiniGPT-Pancreas obtained an accuracy, precision, recall, and F1 score of 0.876, 0.874, 0.878, and 0.876, respectively (Table \ref{tab:tumor_classification}).
The scores of the metrics were compared with the base checkpoint of MiniGPT-v2 and the MiniGPT-Pancreas version for pancreas detection (Table \ref{tab:tumor_classification}). The results have shown that, while there is no noticeable improvement between the base MiniGPT-v2 and MiniGPT-Pancreas for organ detection, the second fine-tuning led to a boost in the metrics, exceeding 0.40 for accuracy and precision, and about 0.50 for recall and F1. This highlights the impact of the task identifier, which was \texttt{VQA} for the classification task, on model performances. 

\begin{table}[ht]
    \centering
    \caption{Tumor classification results. The best results are in bold.}
    \label{tab:tumor_classification}
    \resizebox{0.7\columnwidth}{!}{%
    \begin{tabular}{cccccc}
        \hhline{=====}
        \textbf{Model} & \textbf{Accuracy} & \textbf{Precision}& \textbf{Recall} & \textbf{F1 Score}\\ \hline
        MiniGPT-v2    & 0.462  & 0.444  & 0.304 & 0.361  \\
        \makecell{Ours (Pancreas detection\\fine-tuning)}  & 0.460  & 0.448  & 0.341  & 0.387 \\
        \makecell{Ours (Pancreas tumor\\ classification fine-tuning)} & \textbf{0.876}  & \textbf{0.874}  & \textbf{0.878}  & \textbf{0.876}   \\ 
        \hhline{=====}
    \end{tabular}
    }
\end{table}

\subsection{Pancreas Tumor Detection}
\label{subsec:tumor_detection}

For pancreas tumor detection, the checkpoint of MiniGPT-Pancreas employed for the second task was further fine-tuned. In this case, only the MSD dataset was used, since it includes also subjects with annotations of tumors, as stated in Section \ref{subsec:training_pipeline}.
Compared with the detection task on pancreas parenchyma, the results were much lower, with an IoU of 0.168. This decrease in performance was mainly due to the smaller size of pancreas tumors than the organ itself. 
As shown in Table \ref{tab:tumor_detection}, this final fine-tuning demonstrated a substantial improvement in the average IoU from the previous fine-tuning of MiniGPT-Pancreas and the base MiniGPT-v2.
Some examples of the pancreas tumor detection task on the MSD dataset are shown in Fig. \ref{app-fig:MSD_tumor_detection} of the Appendix.

\begin{table}[ht]
    \centering
    \caption{Results on the tumor detection task. The best results are in bold.}
    \label{tab:tumor_detection}
    \resizebox{.5\columnwidth}{!}{%
    \begin{tabular}{cc}
        \hhline{==}
        \textbf{Model} & \textbf{IoU}  \\ \hline
        MiniGPT-v2    & $ < $ 0.001      \\
        \makecell{Ours (Pancreas detection fine-tuning)}  & 0.025 \\
        \makecell{Ours (Tumor classification fine-tuning)} & 0.072 \\
        \makecell{Ours (Tumor detection fine-tuning)}    & \textbf{0.168} \\
        \hhline{==}
    \end{tabular}
    }
\end{table}

\subsection{Multi-organ Detection}
\label{subsec:multi-organ_detection}

Multi-organ detection was performed using the AbdomenCT-1K dataset with annotations of the liver, spleen, pancreas, and kidneys. The scores on IoU for each organ are presented in Table \ref{tab:multi_organ_evaluation}. As shown, the best-performing organ was the liver, followed by the kidney, the spleen, and, lastly, the pancreas. In particular, the IoU on pancreas detection on a multi-organ dataset like AbdomenCT-1k was approximately 8\% lower than the average value on datasets with single organ annotations like the NIH, and MSD (Table \ref{tab:epoch60}).
Some examples of the multi-organ detection task on the AbdomenCT-1k dataset are shown in Fig. \ref{app-fig:AbdomenCT-1k_detection} of the Appendix.
\begin{table}[h]
    \centering
    \caption{Performance metrics for multi-organ detection on the AbdomenCT-1K dataset.}
    \label{tab:multi_organ_evaluation}
    \resizebox{0.25\columnwidth}{!}{%
    \begin{tabular}{lc}
        \hhline{==}
        \textbf{Organ} & \textbf{IoU}  \\ \hline
        Liver        & 0.839 \\
        Right Kidney & 0.722\\
        Spleen       & 0.705 \\
        Pancreas     & 0.497  \\
        \hline
        Average      & 0.691\\
        \hhline{==}
    \end{tabular}
    }
\end{table}

\subsection{Qualitative Results}
\label{subsec:qualitative_results}
\begin{figure}[h]
    \centering
    \includegraphics[width=1\linewidth]{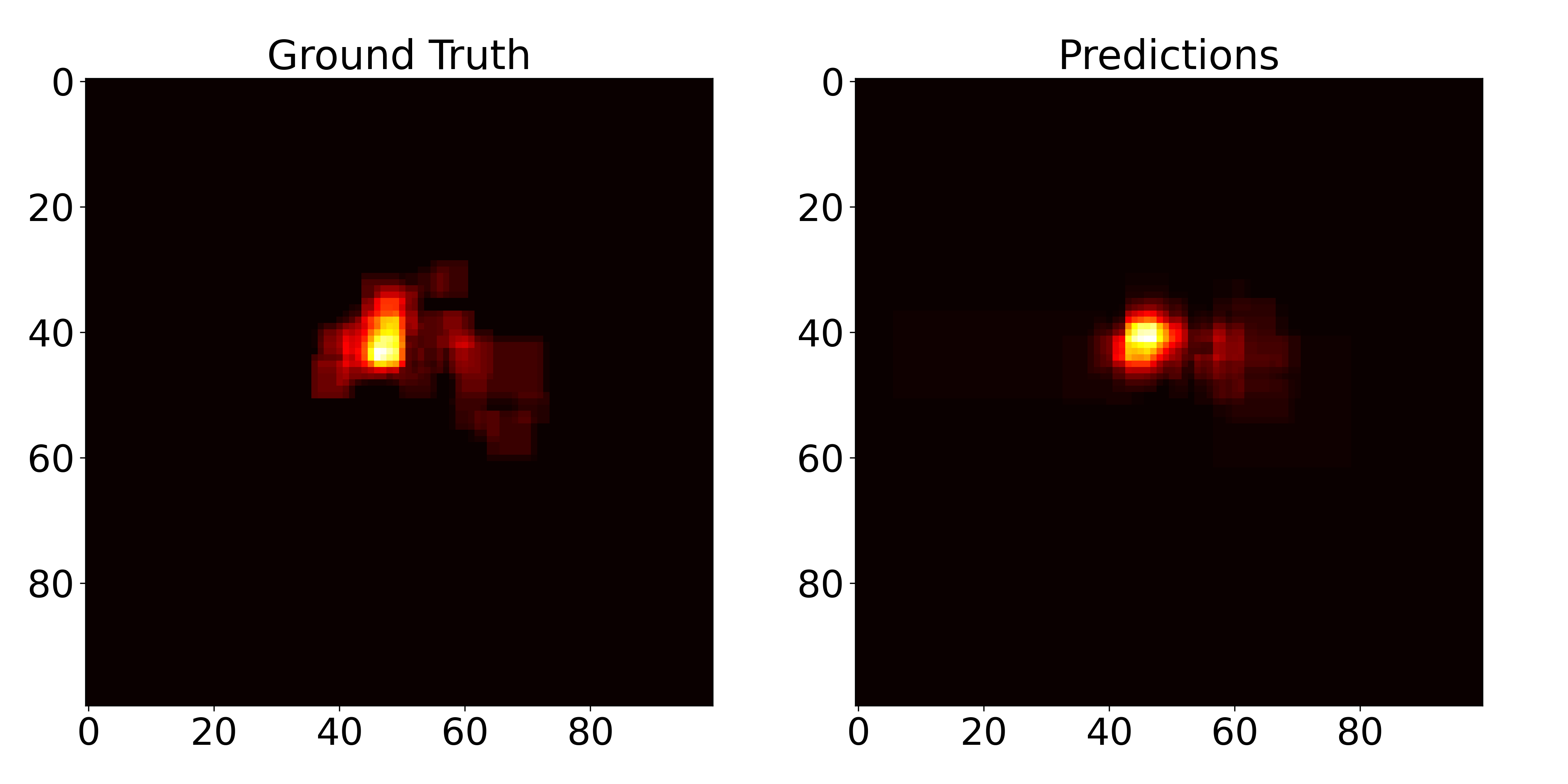}
    \caption{Heatmaps of the GTs and predicted bounding boxes for the pancreas tumor detection task.}
    \label{fig:tumor_heatmap}
\end{figure}

While a quantitative metric such as IoU offered an objective measure of performance, it did not fully capture the model's applicability in clinical settings. The normalized heatmaps of the bounding boxes of the GTs and predictions, depicted in Fig. \ref{fig:tumor_heatmap}, show that the overall position of the predicted bounding boxes was quite similar to the one of the GTs, indicating that the model learned where the pancreas tumor was located.

\section{Discussion}
\label{sec:discussion}

The recent explosion of Generative AI has led to a run-up to increasingly high-performing LLMs that have been extended to the multimodal domain. However, applying MLLMs to pancreas imaging tasks is demanding 
due to the small size, blurred boundaries, and variability of shape, and position of the organ among patients. 

This work provided a proof of concept of MiniGPT-Pancreas, the first MLLM for pancreas tumor diagnosis, combining text and CT scans. It laid its foundation on publicly available resources, like the base MLLM (MiniGPT-v2), and the datasets of radiological images of the pancreas (NIH, MSD, and AbdomenCT-1k).
Thanks to the LoRA technique for fine-tuning, the size of the LLM component was reduced to 0.5\% of the original LLaMA-2-chat-7b-hf model, thus limiting the hardware resources for fine-tuning to one single GPU with 40GB of memory.
To foster continual advancements in this field we made our code available on a GitHub repository (\href{https://github.com/elianastasio/MiniGPT-Pancreas}{https://github.com/elianastasio/MiniGPT-Pancreas}).

Our findings have shown that the cascaded approach of fine-tuning was beneficial in improving considerably the performances of the original MiniGPT-v2 model through various tasks. 
In particular, MiniGPT-Pancreas could accurately distinguish patients with and without pancreas cancer, achieving accuracy, precision, and recall of approximately 0.870 on the NIH and MSD publicly available datasets, equivalent to an improvement of over 0.400 on accuracy and precision, and 0.500 on recall w.r.t. the values achieved by the original MiniGPT-v2. 
For a comparison with deep learning models, convolutional neural networks trained on the National Taiwan
University Hospital, reported an accuracy of 0.830 and a recall of 0.790 when tested for generalization on the combination of the NIH and MSD datasets \citep{liu2020deep}.
The detection task has proved to be tougher, especially for the detection of pancreatic tumors, reporting an IoU of 0.168 in the MSD dataset, much lower than the detection of the pancreas parenchyma, approximately 0.595 and 0.550 on the NIH and MSD datasets, respectively. 
Our findings confirmed the challenges in the detection of small neoplasms using MLLMs, as demonstrated recently by MiniGPT-Med, a fine-tuned version of MiniGPT-v2, reporting an IoU of 0.260 for lung nodules detection \citep{alkhaldi2024minigpt}.
The drop of the IoU can be attributed to the high sensitivity of this metric to small localization errors (quantifiable in a few pixels
in some cases), especially when the GT bounding box is small, as in the case of pancreas tumors. 
Nevertheless, as a first attempt towards alignment with clinical objectives, even a 0.168 IoU can be valuable in directing attention to specific anatomical regions, potentially preventing misjudgment, especially when used by inexperienced radiologists who may have missed initially - and unintentionally - the recognition of a tumor.
Moreover, our findings on both detection tasks represent a significant improvement compared to the base MiniGPT-v2 model. For the pancreas parenchyma detection, MiniGPT-Pancreas improved the IoU of MiniGPT-v2 of more than 11 and 12 times on the NIH and MSD datasets, respectively. For the tumor detection, the improvement exceeded two orders of magnitude.

Our results have highlighted that the fine-tuned model performed better for pancreas detection on datasets with only pancreas annotations than with multi-organ annotations. The observed difference of about 8\% was due to the demanding task of detecting four organs simultaneously. 

MiniGPT-Pancreas may open up new scenarios to supplement clinicians for decision-making and is flexible for several reasons. First, in contrast with other open-source MLLMs like LLaVA-Med and PeFoMed, it is not limited to the VQA, but it can be applied to other tasks, e.g. detection. Second, the results of multi-organ detection have shown that it can be used for the detection of other organs. Third, the bounding box generated by MiniGPT-Pancreas can be used as a prompt to guide foundation models like Segment Anything Model (SAM) for segmentation \citep{mazurowski2023segment}.
However, before the integration with models like SAM, the accuracy on organ detection must be improved.

This study has some limitations.
An MLLM like MiniGPT-Pancreas, processing images one slice at a time, was inherently limited by its inability to take advantage of the 3D spatial coherence. Since the model saw only one 2D slice at a time, it had to infer repeatedly the location of the organ or tumor in each new slice, without any direct reference to its position in previous or subsequent slices. This lack of inter-slice continuity added to the complexity of the task, as MiniGPT-Pancreas had to localize independently the region of interest on each slice. The task resembled detecting an object from scratch in a new image every time, without the benefit of spatial memory that a 3D model enjoys.
This issue largely stemmed from the fact that the ViT employed in the model was designed to process individual 2D images rather than full 3D volumes. As a result, the MiniGPT-Pancreas lacked the contextual information provided by adjacent slices, which could offer critical insights into the spatial relationships and continuity of anatomical structures. To mitigate this issue, only slices with a significant portion of the target were selected, using the bounding box area ratio. Future research work will concern the replacement of the ViT with a 3D visual encoder. Second, the radiological images used for training and testing were only CT slices of the abdominal organs containing the pancreas annotations. Therefore, we could not evaluate the performances on images from different modalities, e.g., MRI or ultrasound, or from different anatomical districts. In a future work we will enlarge the datasets with images from different modalities and organs. Third, MiniGPT-Pancreas was not fine-tuned on tasks like VQA to show the capability to generate open-ended answers. Although the focus of the present work was on the classification and detection of pancreas tumors, we will extend the capabilities of our model to VQA in future work.

\section{Conclusion}
\label{sec:conclusion}

Pancreas tumor is very aggressive and is often diagnosed at an advanced stage. Recognizing the need to improve the accuracy at the early diagnosis stage, many AI models for pancreas imaging have been developed over the past decade. In this work, we presented MiniGPT-Pancreas, the first chatbot for the diagnosis of pancreas tumor after fine-tuning MiniGPT-v2, a general-purpose MLLM. For this specific purpose, a cascaded fine-tuning was performed for the tasks of pancreas detection, tumor classification, and detection on the NIH and MSD publicly available datasets. The results have shown that the MiniGPT-Pancreas obtained a high accuracy, precision, and recall on tumor classification, but it suffered on the detection tasks. Although the IoU on pancreas tumor detection is low, the model can detect tumors in an area close to the actual one, thus supporting young radiologists who may have missed initially the recognition of the cancer.


\clearpage
\appendix
\section*{Appendix}

\section{Materials and Methods}
\label{app-sec:materials_methods}

The following \texttt{keys} were used in the \texttt{JSON} file:

\begin{itemize}
    \item \textbf{\texttt{dataset}}: Refers to the dataset of origin, e.g., the MSD dataset.
    \item \textbf{\texttt{volume\_name}}: The name of the volume file, which contains the slice in question.
    \item \textbf{\texttt{slice\_id}}: A unique identifier for the whole dataset.
    \item \textbf{\texttt{slice\_index}}: The position of this slice within the entire volume.
    \item \textbf{\texttt{slice\_count}}: The total number of slices in the volume.
    \item \textbf{\texttt{pixels\_pancreas}}: The number of pixels labeled as pancreas in the slice. For the MSD dataset, this is the sum of pancreas pixels and tumor pixels.
    \item \textbf{\texttt{pancreas\_pixels\_ratio}}: The ratio of pancreas pixels to the maximum number of pancreas pixels found in any slice of the same volume. A value of 50\%, for example, indicates that the current slice has half the pancreas pixels of the slice with the most pancreas pixels in that volume. This information is necessary for the adopted training approach, described in the next section.
    \item \textbf{\texttt{max\_pixels\_pancreas}}: The maximum number of pancreas pixels found in any slice of the same volume.
    \item \textbf{\texttt{pixels\_tumor}}: The number of pixels identified as tumor in the slice. The information relative to the pancreatic tumor is, of course, present only for the MSD dataset.
    \item \textbf{\texttt{tumor\_pixels\_ratio}}: The ratio of tumor pixels to the maximum number of tumor pixels found in any slice of the same volume.
    \item \textbf{\texttt{max\_pixels\_tumor}}: The maximum number of tumor pixels found in any slice of the same volume.
    \item \textbf{\texttt{bbox\_pancreas}}: The bounding box coordinates for the pancreas in this slice, provided as \textit{(min\_$x$, min\_$y$, max\_$x$, max\_$y$)}.
    \item \textbf{\texttt{pancreas\_bbox\_ratio}}: The ratio of the pancreas bounding box area to the maximum pancreas bounding box area found in any slice of the same volume.
    \item \textbf{\texttt{max\_bbox\_pancreas}}: The maximum area of the pancreas bounding box in any slice of the same volume.
    \item \textbf{\texttt{bbox\_tumor}}: The bounding box coordinates for the tumor in this slice, provided as \textit{(min\_$x$, min\_$y$, max\_$x$, max\_$y$)}.
    \item \textbf{\texttt{tumor\_bbox\_ratio}}: The ratio of the tumor bounding box area to the maximum tumor bounding box area found in any slice of the same volume.
    \item \textbf{\texttt{max\_bbox\_tumor}}: The maximum area of the tumor bounding box in any slice of the same volume.
    \item \textbf{\texttt{width}}: The width of the image in pixels.
    \item \textbf{\texttt{height}}: The height of the image in pixels.
\end{itemize}

\setcounter{figure}{0}
\begin{figure*}[!t]
	\centering
	\includegraphics[width=1.0\linewidth]{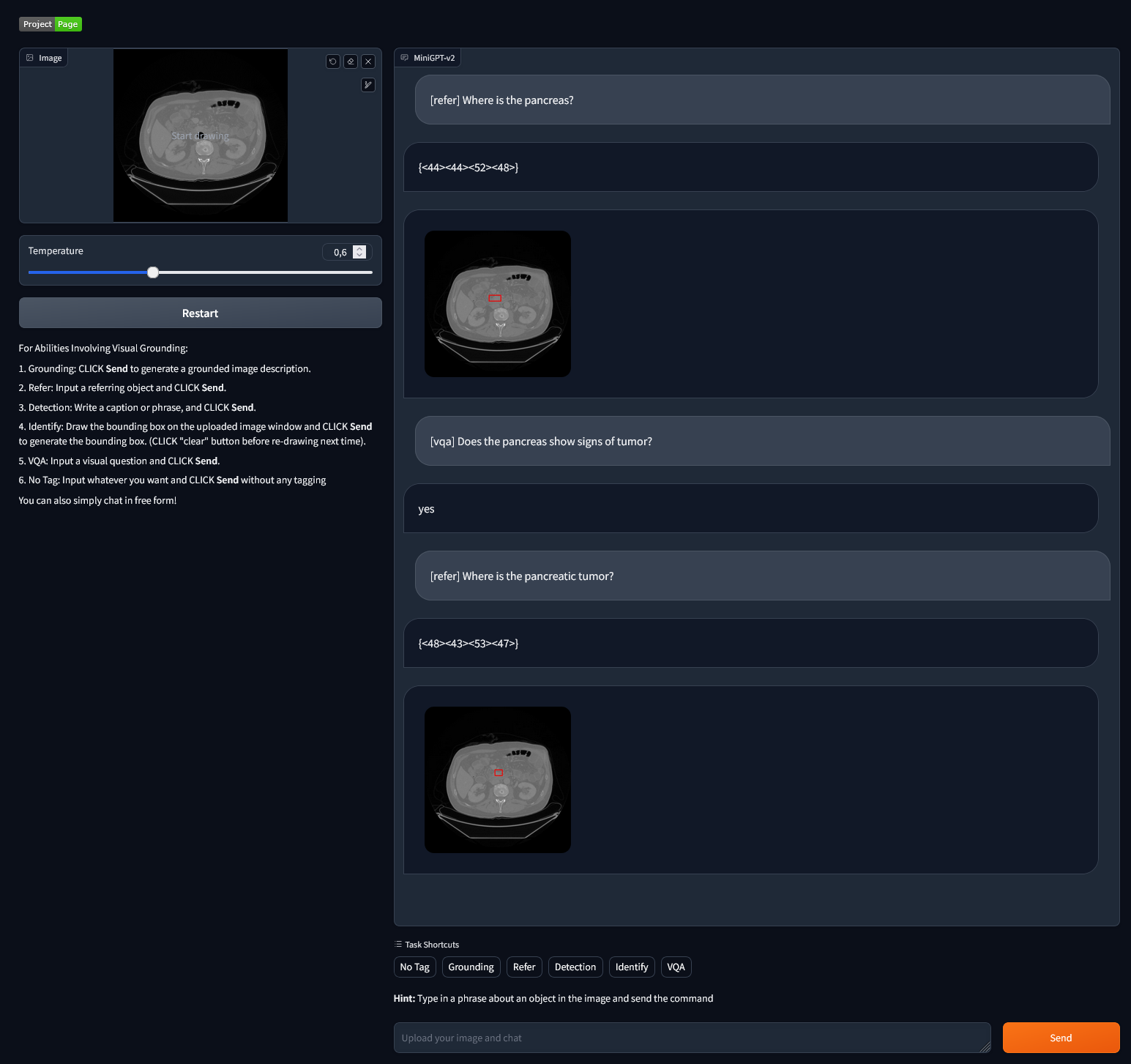}
	\caption{Web interface used for inference, with an example of pancreas detection, tumor classification, and tumor detection. The generated output shows the answer or the coordinates of the bounding box as text. The bounding box is drawn in red on the CT slice.}
	\label{app-fig:minigpt-v2_web_interface}
\end{figure*}

\section{Results}
\label{app-sec:results}

\subsection{Pancreas Detection}
\label{subsec:pancreas_detection}

\setcounter{table}{0}
\begin{table}[h]
    \centering
    \caption{Number of slices in the train and test splits of the NIH and MSD datasets for the pancreas detection task among different thresholds.}
    \label{app-tab:pancreas_detection_sizes}
    \resizebox{0.5\columnwidth}{!}{%
    \scriptsize
    \begin{tabular}{lcccc}
        \hhline{=====}
        \multirow{2}{*}{\textbf{Threshold (\%)}} & \multicolumn{2}{c}{\textbf{NIH}} & \multicolumn{2}{c}{\textbf{MSD}} \\
        \cmidrule(lr){2-5}
        & \textbf{Train} & \textbf{Test} & \textbf{Train} & \textbf{Test} \\
        \midrule
        0   & 5,520 & 1,352 & 6,936 & 1,803 \\
        10  & 4,066 & 1,064 & 5,856 & 1,526 \\
        20  & 2,705 & 720   & 4,391 & 1,149 \\
        30  & 2,180 & 614   & 3,327 & 893   \\
        40  & 1,891 & 544   & 2,710 & 716   \\
        50  & 1,657 & 455   & 2,312 & 621   \\
        60  & 1,402 & 378   & 1,985 & 526   \\
        70  & 1,180 & 306   & 1,681 & 444   \\
        80  & 897  & 226    & 1,389 & 343   \\
        90  & 504  & 135    & 968  & 224    \\
        \hhline{=====}
    \end{tabular}
    }
\end{table}

\begin{table}[h]
    \centering
    \caption{Average IoU for the pancreas detection task}
    \label{app-tab:pancreas_detection_results}
    \resizebox{0.5\columnwidth}{!}{%
    \scriptsize
    \begin{tabular}{lccc}
        \hhline{====}
        \multicolumn{1}{l}{} & \multicolumn{1}{c}{\textbf{NIH}} & \multicolumn{1}{c}{\textbf{MSD}} & \multicolumn{1}{c}{\textbf{Average}} \\
        \cmidrule(){2-4}
        \textbf{Threshold (\%)} & \textbf{IoU} & \textbf{IoU} & \textbf{IoU} \\
        \midrule
        0 & 0.396 & 0.408 & 0.403 \\
        10 & 0.434 & 0.450 & 0.444 \\
        20 & 0.469 & 0.461 & 0.464 \\
        30 & 0.503 & .0474 & 0.486 \\
        40 & 0.529 & 0.500 & 0.512 \\
        50 & 0.550 & 0.510 & 0.527 \\
        60 & 0.587 & 0.541 & 0.560 \\
        70 & 0.606 & 0.577 & 0.589 \\
        80 & 0.626 & 0.591 & 0.605 \\
        90 & 0.634 & 0.599 & 0.612 \\
        \hhline{====}
    \end{tabular}%
    }
\end{table}

\begin{figure*}
  \begin{minipage}[t]{.3\linewidth}
    \includegraphics[width=\linewidth]{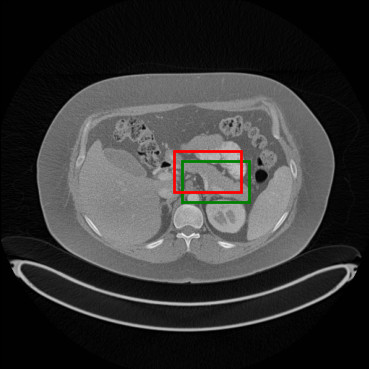}%
    \subcaption{IoU=0.504}%
  \end{minipage}\hfil
  \begin{minipage}[t]{.3\linewidth}
    \includegraphics[width=\linewidth]{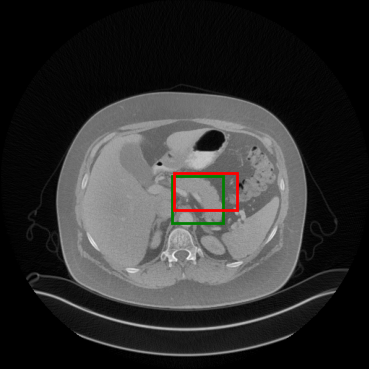}%
    \subcaption{IoU=0.563}%
  \end{minipage}\hfil
  \begin{minipage}[t]{.3\linewidth}
    \includegraphics[width=\linewidth]{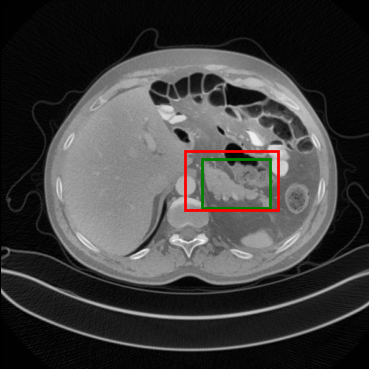}%
    \subcaption{IoU=0.600}%
  \end{minipage}%

  \begin{minipage}[t]{.3\linewidth}
    \includegraphics[width=\linewidth]{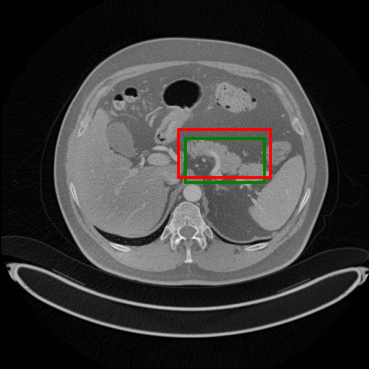}%
    \subcaption{IoU=0.641}%
  \end{minipage}\hfil
  \begin{minipage}[t]{.3\linewidth}
    \includegraphics[width=\linewidth]{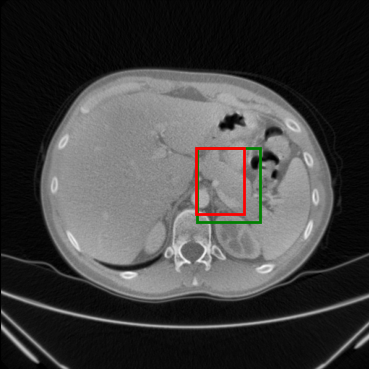}%
    \subcaption{IoU=0.648}%
  \end{minipage}\hfil
  \begin{minipage}[t]{.3\linewidth}
    \includegraphics[width=\linewidth]{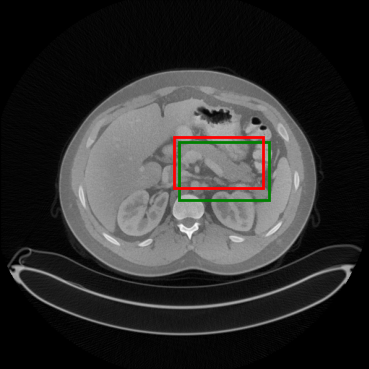}%
    \subcaption{IoU=0.654}%
  \end{minipage}%

  \begin{minipage}[t]{.3\linewidth}
    \includegraphics[width=\linewidth]{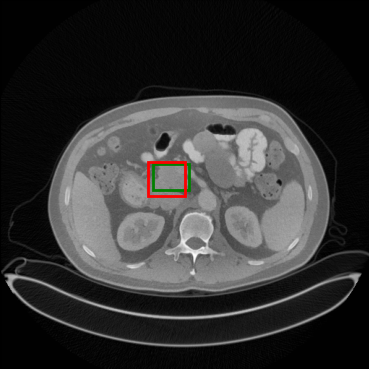}%
    \subcaption{IoU=0.654}%
  \end{minipage}\hfil
  \begin{minipage}[t]{.3\linewidth}
    \includegraphics[width=\linewidth]{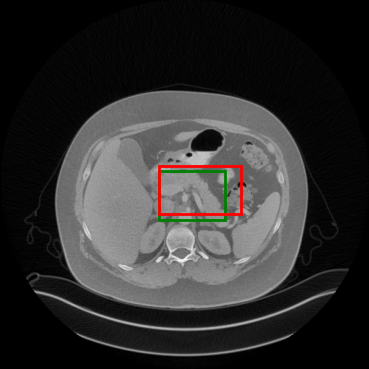}
    \subcaption{IoU=0.661}%
  \end{minipage}\hfil
  \begin{minipage}[t]{.3\linewidth}
    \includegraphics[width=\linewidth]{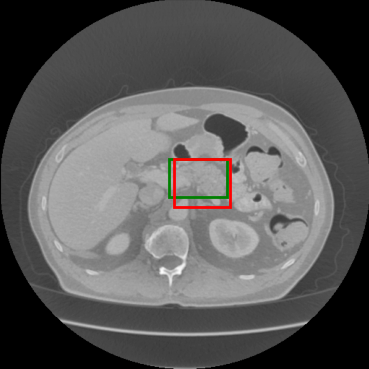}%
    \subcaption{IoU=0.701}%
  \end{minipage}%

  \begin{minipage}[t]{.3\linewidth}
    \includegraphics[width=\linewidth]{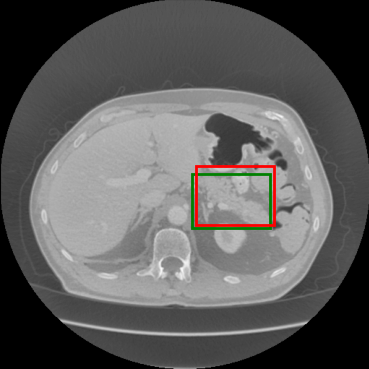}%
    \subcaption{IoU=0.753}%
  \end{minipage}\hfil
  \begin{minipage}[t]{.3\linewidth}
    \includegraphics[width=\linewidth]{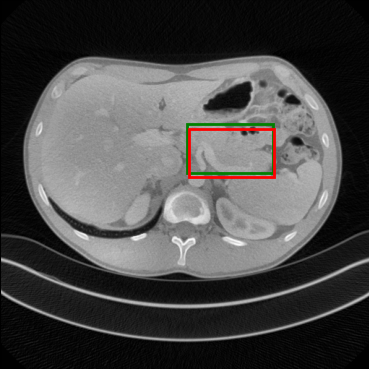}%
    \subcaption{IoU=0.793}%
  \end{minipage}\hfil
  \begin{minipage}[t]{.3\linewidth}
    \includegraphics[width=\linewidth]{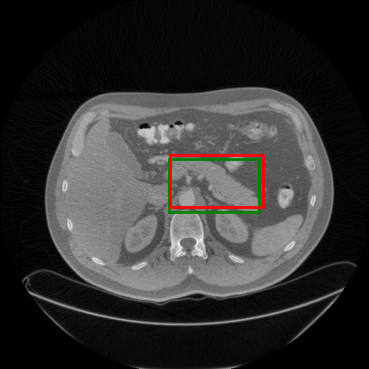}%
    \subcaption{IoU=0.802}%
  \end{minipage}%

    \caption{Examples of the pancreas detection task using the NIH dataset. The green bounding box represents the GT, while the red represents the prediction. The IoU score is displayed below each image.}
    \label{app-fig:NIH_pancreas_detection}
  
\end{figure*}

\begin{figure*}
  \begin{minipage}[t]{.3\linewidth}
    \includegraphics[width=\linewidth]{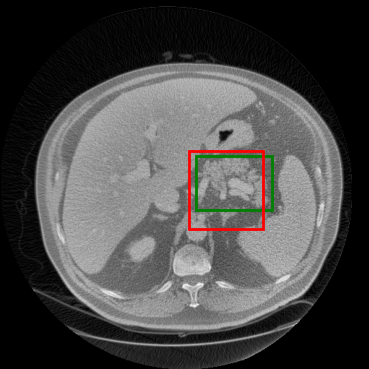}%
    \subcaption{IoU=0.580}%
  \end{minipage}\hfil
  \begin{minipage}[t]{.3\linewidth}
    \includegraphics[width=\linewidth]{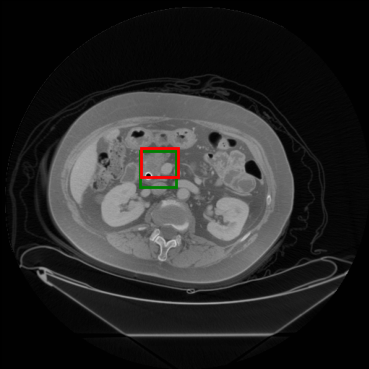}%
    \subcaption{IoU=0.632}%
  \end{minipage}\hfil
  \begin{minipage}[t]{.3\linewidth}
    \includegraphics[width=\linewidth]{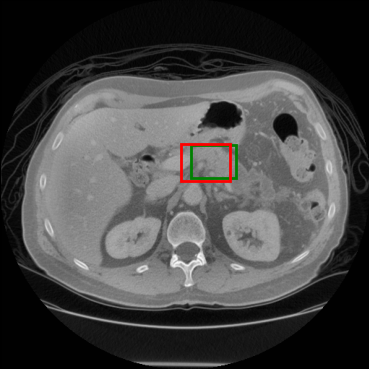}%
    \subcaption{IoU=646}%
  \end{minipage}%

  \begin{minipage}[t]{.3\linewidth}
    \includegraphics[width=\linewidth]{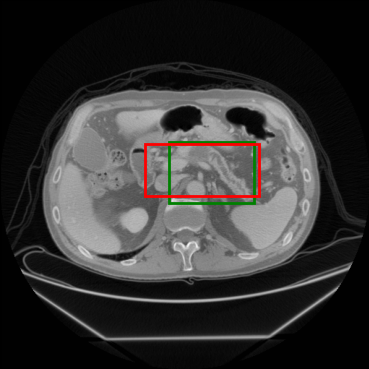}%
    \subcaption{IoU=0.653}%
  \end{minipage}\hfil
  \begin{minipage}[t]{.3\linewidth}
    \includegraphics[width=\linewidth]{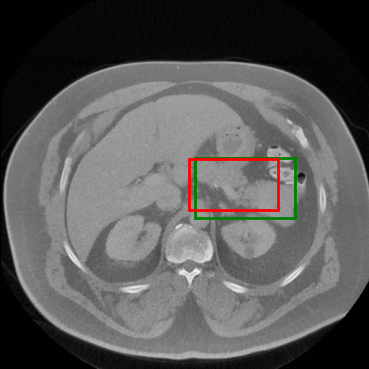}%
    \subcaption{IoU=0.681}%
  \end{minipage}\hfil
  \begin{minipage}[t]{.3\linewidth}
    \includegraphics[width=\linewidth]{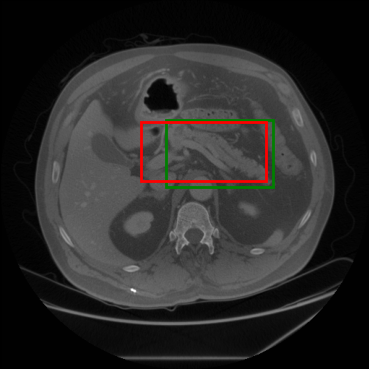}%
    \subcaption{IoU=0.685}%
  \end{minipage}%

  \begin{minipage}[t]{.3\linewidth}
    \includegraphics[width=\linewidth]{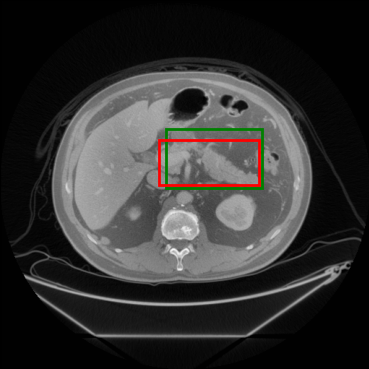}%
    \subcaption{IoU=0.690}%
  \end{minipage}\hfil
  \begin{minipage}[t]{.3\linewidth}
    \includegraphics[width=\linewidth]{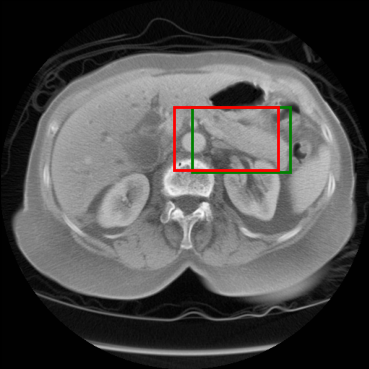}
    \subcaption{IoU=0.715}%
  \end{minipage}\hfil
  \begin{minipage}[t]{.3\linewidth}
    \includegraphics[width=\linewidth]{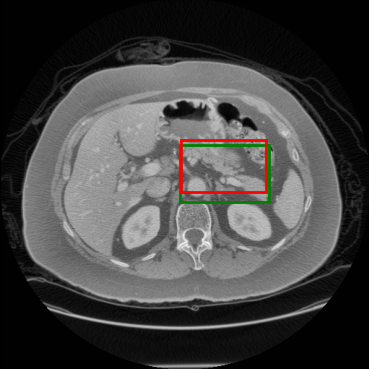}%
    \subcaption{IoU=0.733}%
  \end{minipage}%

  \begin{minipage}[t]{.3\linewidth}
    \includegraphics[width=\linewidth]{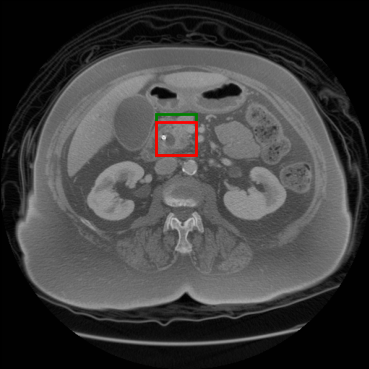}%
    \subcaption{IoU=0.791}%
  \end{minipage}\hfil
  \begin{minipage}[t]{.3\linewidth}
    \includegraphics[width=\linewidth]{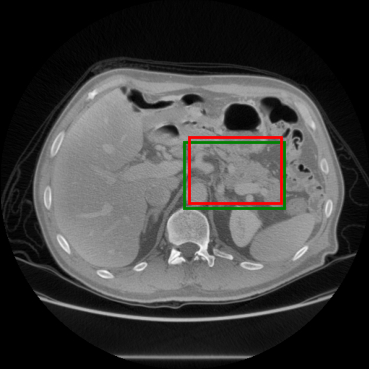}%
    \subcaption{IoU=0.800}%
  \end{minipage}\hfil
  \begin{minipage}[t]{.3\linewidth}
    \includegraphics[width=\linewidth]{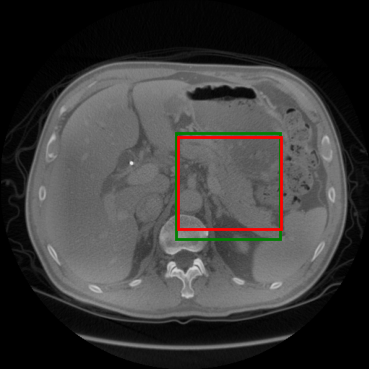}%
    \subcaption{IoU=0.860}%
  \end{minipage}%

    \caption{Examples of the pancreas detection task using the MSD dataset. The green bounding box represents the GT, while the red represents the prediction. The IoU score is displayed below each image.}
    \label{app-fig:MSD_pancreas_detection}
  
\end{figure*}

\begin{figure*}
  \begin{minipage}[t]{.3\linewidth}
    \includegraphics[width=\linewidth]{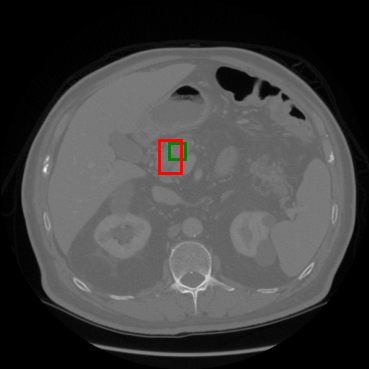}%
    \subcaption{IoU=0.233}%
  \end{minipage}\hfil
  \begin{minipage}[t]{.3\linewidth}
    \includegraphics[width=\linewidth]{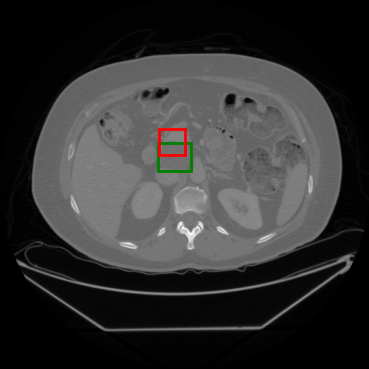}%
    \subcaption{IoU=0.234}%
  \end{minipage}\hfil
  \begin{minipage}[t]{.3\linewidth}
    \includegraphics[width=\linewidth]{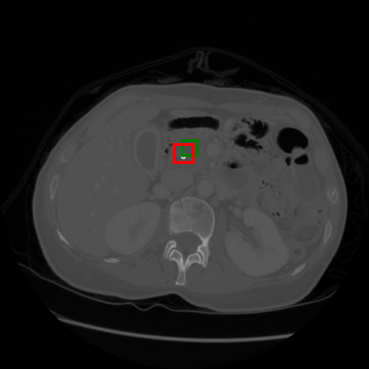}%
    \subcaption{IoU=0.264}%
  \end{minipage}%

  \begin{minipage}[t]{.3\linewidth}
    \includegraphics[width=\linewidth]{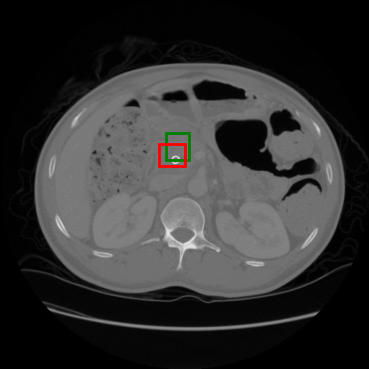}%
    \subcaption{IoU=0.334}%
  \end{minipage}\hfil
  \begin{minipage}[t]{.3\linewidth}
    \includegraphics[width=\linewidth]{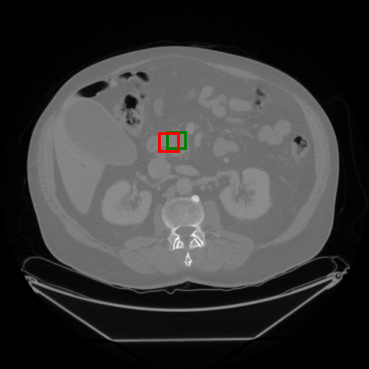}%
    \subcaption{IoU=0.339}%
  \end{minipage}\hfil
  \begin{minipage}[t]{.3\linewidth}
    \includegraphics[width=\linewidth]{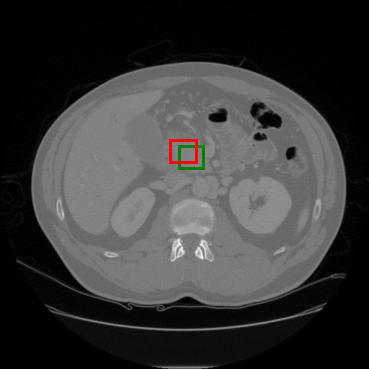}%
    \subcaption{IoU=0.344}%
  \end{minipage}%

  \begin{minipage}[t]{.3\linewidth}
    \includegraphics[width=\linewidth]{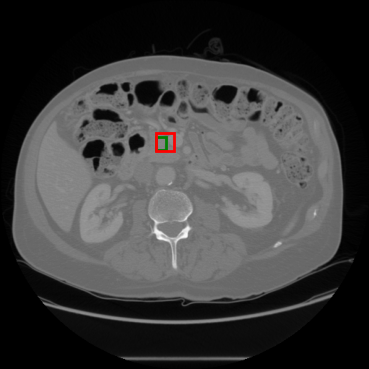}%
    \subcaption{IoU=0.410}%
  \end{minipage}\hfil
  \begin{minipage}[t]{.3\linewidth}
    \includegraphics[width=\linewidth]{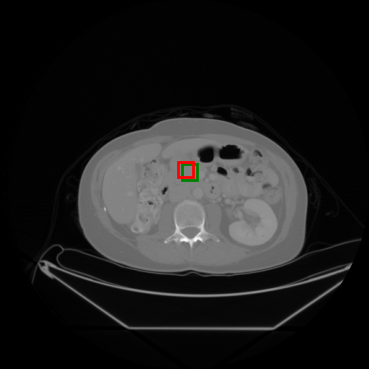}
    \subcaption{IoU=0.417}%
  \end{minipage}\hfil
  \begin{minipage}[t]{.3\linewidth}
    \includegraphics[width=\linewidth]{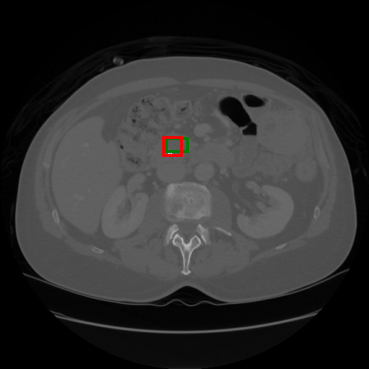}%
    \subcaption{IoU=0.469}%
  \end{minipage}%

  \begin{minipage}[t]{.3\linewidth}
    \includegraphics[width=\linewidth]{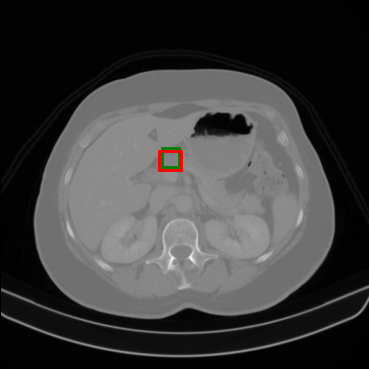}%
    \subcaption{IoU=0.539}%
  \end{minipage}\hfil
  \begin{minipage}[t]{.3\linewidth}
    \includegraphics[width=\linewidth]{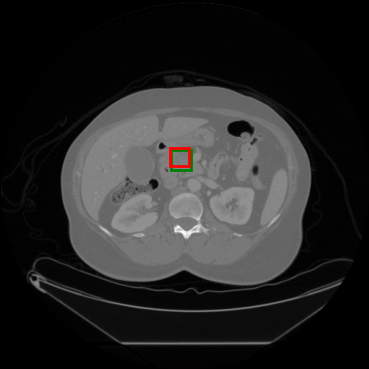}%
    \subcaption{IoU=0.620}%
  \end{minipage}\hfil
  \begin{minipage}[t]{.3\linewidth}
    \includegraphics[width=\linewidth]{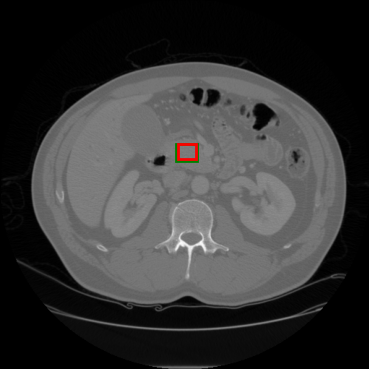}%
    \subcaption{IoU=0.739}%
  \end{minipage}%

    \caption{Examples of the pancreas tumor detection task using the MSD dataset. The green bounding box represents the GT, while the red represents the prediction. The IoU score is displayed below each image.}
    \label{app-fig:MSD_tumor_detection}
  
\end{figure*}

\begin{figure*}
  \begin{minipage}[t]{.3\linewidth}
    \includegraphics[width=\linewidth]{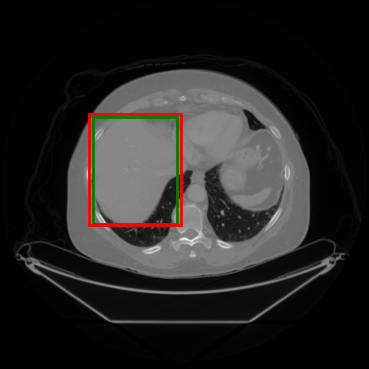}%
    \subcaption{Liver, IoU=0.867}%
  \end{minipage}\hfil
  \begin{minipage}[t]{.3\linewidth}
    \includegraphics[width=\linewidth]{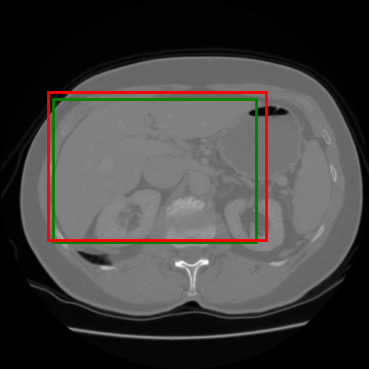}%
    \subcaption{Liver, IoU=0.876}%
  \end{minipage}\hfil
  \begin{minipage}[t]{.3\linewidth}
    \includegraphics[width=\linewidth]{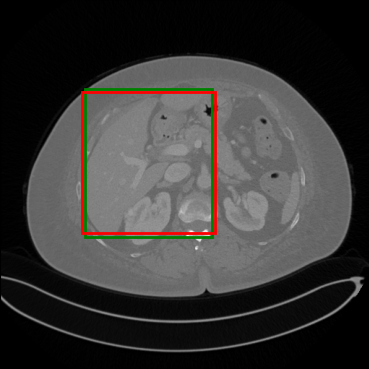}%
    \subcaption{Liver, IoU=0.904}%
  \end{minipage}%

  \begin{minipage}[t]{.3\linewidth}
    \includegraphics[width=\linewidth]{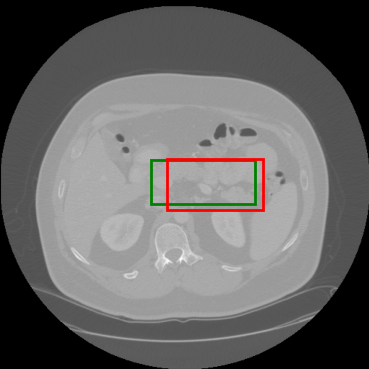}%
    \subcaption{Pancreas, IoU=0.689}%
  \end{minipage}\hfil
  \begin{minipage}[t]{.3\linewidth}
    \includegraphics[width=\linewidth]{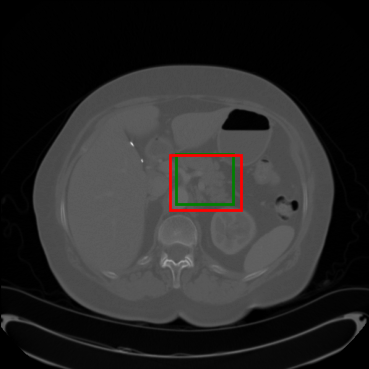}%
    \subcaption{Pancreas, IoU=0.702}%
  \end{minipage}\hfil
  \begin{minipage}[t]{.3\linewidth}
    \includegraphics[width=\linewidth]{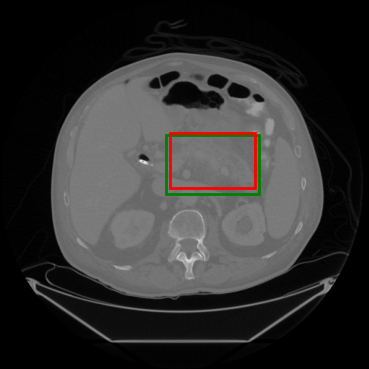}%
    \subcaption{Pancreas, IoU=0.801}%
  \end{minipage}%

  \begin{minipage}[t]{.3\linewidth}
    \includegraphics[width=\linewidth]{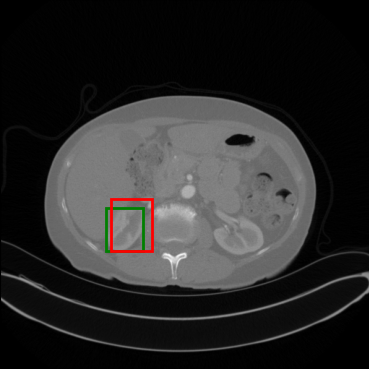}%
    \subcaption{Kidney, IoU=0.601}%
  \end{minipage}\hfil
  \begin{minipage}[t]{.3\linewidth}
    \includegraphics[width=\linewidth]{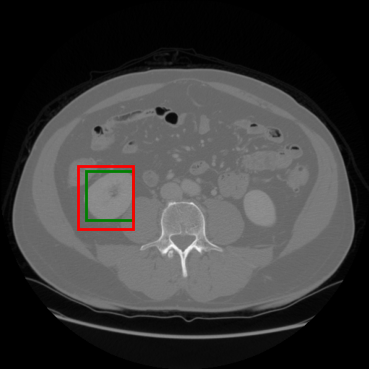}
    \subcaption{Kidney, IoU=0.671}%
  \end{minipage}\hfil
  \begin{minipage}[t]{.3\linewidth}
    \includegraphics[width=\linewidth]{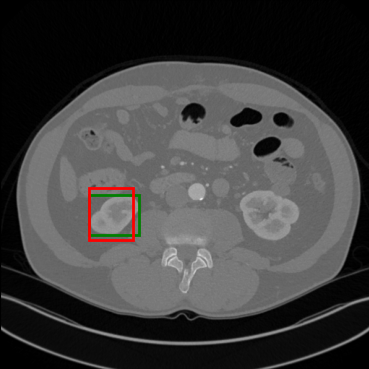}%
    \subcaption{Kidney, IoU=0.699}%
  \end{minipage}%

  \begin{minipage}[t]{.3\linewidth}
    \includegraphics[width=\linewidth]{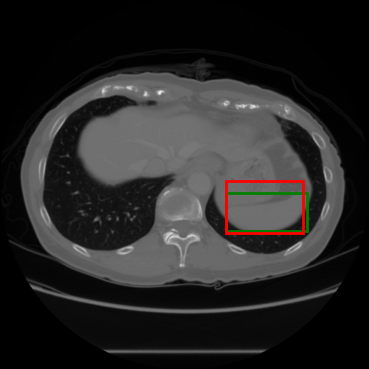}%
    \subcaption{Spleen, IoU=0.722}%
  \end{minipage}\hfil
  \begin{minipage}[t]{.3\linewidth}
    \includegraphics[width=\linewidth]{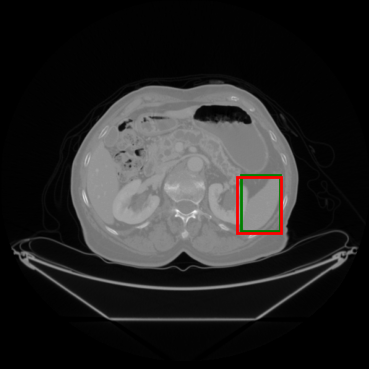}%
    \subcaption{Spleen, IoU=0.835}%
  \end{minipage}\hfil
  \begin{minipage}[t]{.3\linewidth}
    \includegraphics[width=\linewidth]{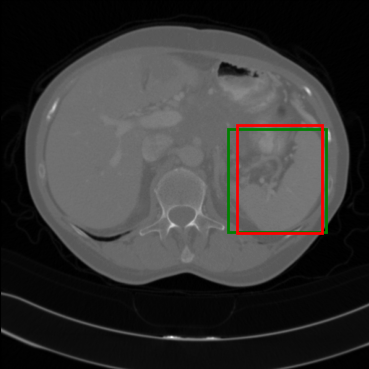}%
    \subcaption{Spleen, IoU=0.844}%
  \end{minipage}%

    \caption{Examples of multi-organ detection task using the AbdomenCT-1k dataset. The green bounding box represents the GT, while the red represents the prediction. The IoU score is displayed below each image.}
    \label{app-fig:AbdomenCT-1k_detection}
  
\end{figure*}

\end{document}